\documentclass{article}

\usepackage{microtype}
\usepackage{graphicx}
\usepackage{subfigure}
\usepackage{booktabs} 
\usepackage[nohyperref, accepted]{icml2021}
\usepackage[hidelinks]{hyperref}
\usepackage{amsmath, amssymb, amsfonts, amsthm, bm}
\usepackage{neuralnetwork}

\icmltitlerunning{An Overview of Uncertainty Quantification Methods for Infinite Neural Networks}

\begin{document}

\twocolumn[
\icmltitle{An Overview of Uncertainty Quantification Methods\\for Infinite Neural Networks}

\icmlsetsymbol{equal}{*}

\begin{icmlauthorlist}
\icmlauthor{Florian Juengermann}{seas}
\icmlauthor{Maxime Laasri}{seas}
\icmlauthor{Marius Merkle}{seas}\\ \ \\
\emph{Supervised by:}
\icmlauthor{Weiwei Pan}{seas}
\end{icmlauthorlist}

\icmlaffiliation{seas}{John A. Paulson School of Engineering and Applied Sciences, Harvard University, Cambridge, MA, USA}

\icmlkeywords{Machine Learning, Uncertainty, Infinite Neural Networks}

\vskip 0.3in
]



\printAffiliationsAndNotice{}

\begin{abstract}

To better understand the theoretical behavior of large neural networks, several works have analyzed the case where a network's width tends to infinity. In this regime, the effect of random initialization and the process of training a neural network can be formally expressed with analytical tools like Gaussian processes and neural tangent kernels. In this paper, we review methods for quantifying uncertainty in such infinite-width neural networks and compare their relationship to Gaussian processes in the Bayesian inference framework. We make use of several equivalence results along the way to obtain exact closed-form solutions for predictive uncertainty.

\end{abstract}

\section{Introduction}
Deep learning has brought breakthroughs across a variety of disciplines including image classification, speech recognition, and natural language processing \cite{lecun2015deep}. But despite their tremendous success, artificial neural networks often tend to be overconfident in their predictions, especially for out-of-distribution data.

Therefore, proper quantification of uncertainty is of fundamental importance to the deep learning community. It turns out that the infinite-width limit, i.e. the limit when the number of hidden units in a neural network approaches infinity, simplifies the analysis considerably, and is useful to draw conclusions for very large, but finite neural networks.

In this paper, we provide an overview of the most common techniques for quantifying uncertainty in deep learning and explore how these techniques are related in the infinite-width regime. In \autoref{section:background}, we introduce the fundamentals of these techniques, while in \autoref{section:infinitewidth}, we outline their relation as well as several equivalence results.

\section{Background}
\label{section:background}

Throughout this section, we will consider a neural network with input $\bm x \in \mathbb{R}^n$ and scalar output $y \in \mathbb{R}$. Let $f(\bm x, \bm \theta)$ denote the neural network output function where the  weights and biases are summarized in the parameter vector $\bm \theta$. The traditional training objective is to minimize the mean squared error (MSE) loss $\mathcal L$ over a dataset $\mathcal{D} = (\mathcal{X}, \mathcal{Y}) = \{(\bm x_i, y_i)\}_{i = 1}^N$ of $N$ training points, i.e. to find an optimal $\bm \theta^*$ such that:
\begin{equation}
    \bm \theta^* = \text{arg min}_{\bm \theta} \underbrace{\frac 1N \sum_{i = 1}^N \left(y_i - f(\bm x_i, \bm \theta) \right)^2}_{\mathcal{L} (\mathcal{X}, \mathcal{Y})}
\end{equation}

\subsection{Gaussian Processes}

One fundamental tool used in machine learning to quantify uncertainty is Gaussian processes (GPs). They are primarily encountered in regression problems, but they can be extended to classification or clustering tasks \cite{kapoor2010gaussian, kim2007clustering}.

Simply put, a GP works as follows: (i) for any input $\bm x \in \mathbb{R}^n$, it yields a probability distribution for any possible outcome $y \in \mathbb{R}$, and (ii) any finite collection of input points follows a multivariate Gaussian distribution $\mathcal{N}(\underline \mu, \underline \Sigma)$ with mean vector $\underline \mu \in \mathbb{R}^n$ and covariance matrix $\underline \Sigma \in \mathbb{R}^{n \times n}$ where one usually assumes the data to be centered ($\underline \mu = 0$). The matrix $\underline \Sigma$ is determined by a kernel function $\Sigma$ that allows to incorporate prior knowledge about the shape and characteristics of the data distribution. We will denote a GP with kernel function $\Sigma$ applied on centered data as $\mathcal{GP}(0,\Sigma)$. The most commonly used kernel function is the squared exponential or radial basis function (RBF):
\begin{equation}
    \underline \Sigma_{ij}= \Sigma(\bm x_i, \bm x_j) = \sigma^2 \exp \left( \frac{|| \bm x_i - \bm x_j||_2^2}{2l^2} \right)
    \label{eq:RBF}
\end{equation}

where the variance $\sigma$ and length $l$ serve as hyperparameters to encode domain knowledge. For a new test point $\bm x$, the prior distribution over $y$ is given by:
\begin{equation}
    y \sim \mathcal{N} \left(0, \Sigma(\bm x, \bm x) \right)
    \label{eq:priordistribution}
\end{equation}

Then, the dataset $\mathcal{D}$ can be used as part of the Bayesian framework to update prior beliefs and obtain a posterior distribution $P$ by conditioning on the training points, as follows:

\begin{equation}
    \begin{split}
    P(&y | \bm x, \mathcal{X}, \mathcal{Y}) = \mathcal{N} \big( \Sigma(\bm x, \mathcal{X}) \Sigma(\mathcal{X}, \mathcal{X})^{-1} \mathcal{Y},  \\
    &\Sigma(\bm x, \bm x) -  \Sigma(\bm x, \mathcal{X})\Sigma(\mathcal{X}, \mathcal{X})^{-1}\Sigma(X, \bm x) \big)
    \end{split}
    \label{eq:gp_posteriordistribution}
\end{equation}

\subsection{Frequentist versus Bayesian Approach}

The most common tools used to quantify uncertainty specifically in neural networks come from the distinction between two fundamental approaches. On the one hand, we can use the variance of predictions of multiple neural networks that follow from different random initializations. We refer to this frequentist type of ensemble method as deep ensemble (DE). On the other hand, a Bayesian neural network (BNN) is the natural result of applying Bayesian inference to neural networks. In a BNN, we place priors -- typically independent Gaussians -- on the network weights and biases. As a result, the posterior predictive distribution gives a prediction interval instead of a single-point estimate.

BNNs used in practice have posterior distributions that are highly complex and multi-modal, challenging even the most sophisticated Markov chain Monte Carlo (MCMC) sampling methods. For that reason, practitioners oftentimes use mean-field variational inference (MFVI) to approximate the posterior with a factorized Gaussian, i.e. assume the posterior distributions over the network's parameters to be independent.

\subsection{Neural Linear Models}
An alternative to MFVI that scales better to large datasets than BNNs and GPs are neural linear models (NLMs).

\begin{figure}
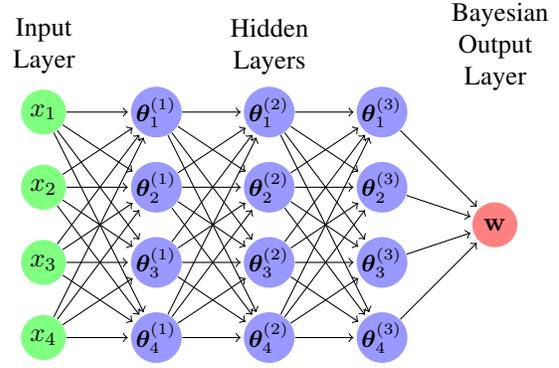

    \centering
    \begin{neuralnetwork}[height=4, layerspacing=15mm]
        \newcommand{\x}[2]{$x_#2$}
        \newcommand{\tfirst}[2]{\small $\bm \theta^{(1)}_{#2}$}
        \newcommand{\tsecond}[2]{\small $\bm \theta^{(2)}_{#2}$}
        \newcommand{\tthird}[2]{\small $\bm \theta^{(3)}_{#2}$}
        \newcommand{\w}[2]{$\mathbf{w}$}
        \inputlayer[count=4, bias=false, title=Input\\Layer, text=\x]
        \hiddenlayer[count=4, bias=false, text=\tfirst] \linklayers
        \hiddenlayer[count=4, bias=false, title=Hidden\\Layers, text=\tsecond] \linklayers
        \hiddenlayer[count=4, bias=false, text=\tthird] \linklayers
        \outputlayer[count=1, bias=false, title= Bayesian\\Output\\Layer, text=\w] \linklayers
    \end{neuralnetwork}
    \caption{Example of a neural linear model. Hidden layer $l$ has vector weights $\bm \theta_i^{(l)}$, while the output layer performs Bayesian linear regression on the weights $\mathbf{w}$.}
    \label{fig:nlm}
\end{figure}

First introduced by \citet{lazaro2010marginalized} as Marginalized Neural Networks and later re-introduced by \citet{snoek2015scalable}, NLMs consist of a Bayesian linear regression performed on the last layer of a deep neural network. They resemble BNNs in structure, except only the final linear layer has priors on its coefficients. An example is shown in \autoref{fig:nlm}.

The first layers whose weights $\bm \theta$ are simple scalars can be viewed as a feature map $\Phi_{\bm \theta}$ that transforms input $\bm x$ into features $\Phi_{\bm \theta}(\bm x)$, which are then fed into the Bayesian linear regression. Because of this particular structure, training and inference with NLMs differs from usual procedures and usually consist of two steps: training the deterministic layers first using a point estimate of the output layer weights, and computing the posterior of that Bayesian layer second. In \cite{thakur2021uncertaintyaware}, the authors propose an instance of this procedure that has the benefit of not underestimating uncertainty in data-scarce regions.

When all priors on the output weights $\mathbf{w}$ are independent and identical Gaussian distributions, the posterior is a multivariate Gaussian whose mean $\underline \mu_{NLM}$ and covariance matrix $\underline \Sigma_{NLM}$ can be expressed in closed-form:
\begin{align}
    {\underline \mu}_{NLM} &=\alpha\left(\alpha{\Phi_{\bm{\theta}}(\mathcal{X})}^{\top}\Phi_{\bm{\theta}}(\mathcal{X})+\beta I\right)^{-1}{\Phi_{\bm{\theta}}(\mathcal{X})}^{\top}\mathcal{Y}\\
    {\underline \Sigma}_{NLM} &=\left(\alpha{\Phi_{\bm{\theta}}(\mathcal{X})}^{\top}\Phi_{\bm{\theta}}(\mathcal{X})+\beta I\right)^{-1}
\label{eq:nlmposterior}
\end{align}

where $I$ is the identity matrix, and $\alpha = \frac 1 {{\sigma_\epsilon}^2}$ and $\beta = \frac 1 {{\sigma_\mathbf{w}}^2}$ are prior hyperparameters equal to the inverse of the variance of the regression noise and weight variables, respectively. These expressions reveal that the main computational effort of inference with NLMs lies in the inversion of the matrix $\alpha{\Phi_{\bm \theta}(\mathcal{X})}^{T}\Phi_{\bm \theta}(\mathcal{X})+\beta I$, whose dimension is the number of neurons in the last hidden layer. Compared with GPs who scale with the size of the training data set, as explained later in \autoref{subsubsection:gpreg}, NLMs scale with the width of the neural network's last layer, which is often of lower dimensionality. This is where NLMs gain their advantage in tractability and computational efficiency, which is why NLMs have gained popularity.

\subsection{Neural Tangent Kernel}

The last piece of background to understand is the theory describing the training process of a neural network with gradient descent. A helpful tool for that is the neural tangent kernel (NTK) introduced by \citet{jacot2018neural}:
\begin{equation}
    \hat \Theta_t(\bm x, \bm {x'}) = \nabla_{\bm \theta} f(\bm x, \bm {\theta_t})^\top \nabla_{\bm \theta} f(\bm {x'}, \bm {\theta_t})
\end{equation}

In general, this kernel function depends on the random initialization $\bm {\theta_0}$ and changes over time. However, \citet{jacot2018neural} show that in the infinite-width limit, the NTK at initialization $\hat \Theta_0$ becomes deterministic and only depends on the activation function and network architecture. In this case, the NTK also stays constant during training for small enough learning rates: $\hat \Theta_t = \hat \Theta_0$. We call this deterministic and constant kernel $\Theta$. \citet{yang2020tensor} extends these convergence results from multilayer perceptrons (MLPs) to other architectures including convolutional and batch-normalization layers.

Now, we investigate how the NTK describes change in the output function of a NN during training. \citet{jacot2018neural} and \citet{lee2019wide} show that for small enough learning rates, the NN prediction $f_t(\bm x) := f(\bm x, \bm \theta_t)$ at training time $t$ follows an ordinary differential equation (ODE) that is characterized by the NTK:
\begin{equation}
    \frac{\text{d} f_t(\bm x)}{\text{d} t}
    = - \nu \hat \Theta_t(\bm x, \mathcal{X}) \nabla_{f_t(\mathcal{X})} \mathcal{L}
    \label{eq:diffeqNTK}
\end{equation}
Here, $\nu$ denotes the learning rate and $\nabla_{f_t(\mathcal{X})}\mathcal{L}$ describes the gradient of the loss with respect to model's output on the training set. Using an MSE loss for $\mathcal{L}$ and the fact that the NTK stays constant over time, \autoref{eq:diffeqNTK} simplifies to a linear ordinary differential equation with known exponentially decaying solution. So, the output for the training dataset at time $t$ is
\begin{equation}
    f_t(\mathcal{X}) = \mathcal{Y} + \left(f_0(\mathcal{X}) - \mathcal{Y}\right) e^{-\nu \Theta(\mathcal{X}, \mathcal{X})t}
\end{equation}
This shows that the NN achieves zero training loss because the NTK is a positive definite matrix, hence the exponential term converges to zero.

Equivalently, based on the results of \cite{lee2019wide}, \citet{he2020bayesian} show that $f_t$ can be expressed based on the random initialization $f_0$. For $t\rightarrow \infty$ this gives:
\begin{equation}
    f_\infty (\bm x) = f_0(\bm x) - \Theta(\bm x, \mathcal{X})\Theta(\mathcal{X, X})^{-1}(f_0(\mathcal{X}) - \mathcal{Y}) \label{eq:f_linear}
\end{equation}


\section{Behavior at Infinite Width}
\label{section:infinitewidth}

In this section, we summarize the behavior of NNs in the infinite-width limit, i.e. we are interested in the limit as the number of hidden units goes to infinity.

\subsection{Priors in Infinite Neural Networks}
Under the condition that the network parameters follow independent and identical (iid) distributions with zero mean and finite variance and the activation function of the last layer is bounded, the prior predictive distribution converges to a GP with zero mean and finite variance \cite{neal1996priors}. The intuition is as follows: 
for a fixed input $\bm x$, every hidden layer output $\phi_i$ is a random variable with finite mean and variance as the activation function is bounded. When multiplied by the final layer weights, $\phi_i w_i$ are iid random variables with zero mean and finite variance. According to the central limit theorem, their sum converges to a Gaussian distribution. If we rescale the variances according to the network width, the network output becomes a standard normal distribution $f(\bm x) \sim \mathcal{N}(0, 1)$. As this holds true for every $\bm x$, the neural network is equivalent to a GP called the neural network Gaussian process (NNGP). A sampled function from this NNGP is equivalent to the output function of a randomly initialized NN and to a prior predictive sample of a BNN.

While \citet{neal1996priors} gives a proof of existence, i.e. that a NN converges to a GP in the infinite-width limit, Neal does not provide any analytic form of the covariance function $\mathcal{K}$. \citet{williams1997computing} builds on that work and provides closed-form kernel functions for different sets of activation functions and prior distributions on the network parameters.

\subsection{Bayesian Inference}
So far we have not considered any data but made predictions based on randomly drawn initial parameters. In the following sections, we discuss three approaches to incorporate the training data. First, we update the GP prior kernel by conditioning on the training data to obtain a GP posterior kernel. Second, we use gradient-based methods to train the NN. Third, we make the neural network Bayesian and approximate the BNN posterior.

\subsubsection{Gaussian Process Regression}
\label{subsubsection:gpreg}
To go from the prior to the posterior distribution in a GP, we condition the NNGP kernel $\mathcal{K}$ on the dataset $(\mathcal{X, Y})$. Following the Bayesian framework, the result is the posterior GP as defined in \autoref{eq:gp_posteriordistribution}.

A major problem of GPs is their computational effort which is cubic in the number of training points $\mathcal{O}(N^3)$. Common methods have enabled exact inference for a maximum of a few thousand training points only. This makes GPs unsuitable for many real-world applications that require very large datasets to be processed. Even though exact inference on GPs has recently been scaled to a million data points by GPU parallelization \cite{wang2019exact}, NNs provide a more accessible framework in the big data regime.

\subsubsection{Neural Network Training}
NNs are trained with gradient-based optimizers such as full-batch gradient descent or its extensions such as stochastic gradient descent and Adam optimizer. Here, we consider two training methods with different theoretical conclusions.

\paragraph{Weakly Trained NN}
First, we only train the last layer of our NN. This means all other layers act as random feature extractors that are unchanged during training. \citet{lee2019wide} show that for the MSE loss function and a sufficiently small learning rate, the network output during training is an interpolation between the GP prior and GP posterior, and asymptotically converges to the posterior of the NNGP. This means, training only the last layer of a randomly initialized NN is equivalent to sampling a function from the GP with the NNGP prior kernel $\mathcal{K}$ conditioned on the dataset.

\paragraph{Fully Trained NN}
If we use the gradient descent algorithm on the entire network, the NTK $\Theta$ describes the training behavior. After convergence to zero loss, \autoref{eq:f_linear} describes the learned function $f_\infty$.
As $f_0$ can be regarded as a sample from the NNGP $\mathcal{GP}(0, \mathcal{K})$, we write $f_\infty \sim \mathcal{GP}(\mu_{\mathrm{DE}}(\bm x), \Sigma_{\mathrm{DE}}(\bm x, \bm x'))$ with \footnote{The notation $+ h.c.$ means "plus Hermitian conjugate", like in \cite{lee2019wide}}
\begin{align}
\mu_{\mathrm{DE}}(\bm x) &= \Theta_{\bm x \mathcal{X}} \Theta_{\mathcal{XX}}^{-1}\mathcal{Y}\\
\Sigma_{\mathrm{DE}}(\bm x, \bm x') &= \mathcal{K}_{\bm{xx'}} + \Theta_{\bm x \mathcal{X}} \Theta_{\mathcal{XX}}^{-1}\mathcal{K}_{\mathcal{XX}}\Theta_{\mathcal{XX}}^{-1}\Theta_{\mathcal{X}\bm x} \nonumber\\
    &\qquad - (\Theta_{\bm x \mathcal{X}}\Theta_{\mathcal{XX}}^{-1}\mathcal{K}_{\mathcal{X}\bm x'} + h.c.).
\end{align}
where we used subscripts for the kernel arguments in interest of space.
\citet{lee2019wide} observed that this deep ensemble GP does not correspond to a proper Bayesian posterior. This means, in contrast to the weakly trained NNs, random initialization followed by gradient descent training does not give valid posterior predictive function samples. Hence, an ensemble of NNs does not accurately approximate the posterior. 

\paragraph{Bayesian Deep Ensemble}
To address this, \citet{he2020bayesian} propose a slight modification to the gradient descent training. With that, they arrive at the what they call neural tangent kernel Gaussian process (NTKGP) with
\begin{align}
\mu_{\mathrm{NTKGP}}(\bm x) &=   
\Theta_{\bm x\mathcal{X}} \Theta_{\mathcal{XX}}^{-1} \mathcal Y\\
\Sigma_{\mathrm{NTKGP}}(\bm x, \bm x') &= 
\Theta_{\bm{xx'}} - \Theta_{\bm x\mathcal{X}} \Theta_{\mathcal{XX}}^{-1} \Theta_{\mathcal{X}\bm x'}
\end{align}
Note that this corresponds to a GP posterior as defined in \autoref{eq:gp_posteriordistribution}. In comparison to the NNGP posterior we obtain when conditioning the NNGP prior on the data, or when only training the last layer of the NN, here, the prior covariance function is not the NNGP kernel $\mathcal{K}$ but the NTK $\Theta$.

\subsubsection{BNN Posterior Approximation}
Instead of using an ensemble of randomly initialized NNs, we can use a prior distribution on the network weights to obtain a BNN. However, approximation methods struggle with accurately representing the complex BNN posterior. \citet{coker2021wide} show that in the infinite-width limit, the commonly used MFVI approximation fails to learn the data. Specifically, the posterior predictive mean for any input converges to zero, regardless of the input data. In their proof, they assume that the Kullback–Leibler (KL) divergence and the $erf$ activation function are used, but give empirical evidence for $\tanh$ and $ReLU$ activations.

\section{Conclusion}
This work provides an overview of the different methods used for quantifying uncertainty in infinite neural networks, and shows how to obtain analytic expressions for both prior and posterior predictive distributions for that purpose.

While the prior predictive can simply be modeled as a GP, we have outlined three ways to obtain proper posterior predictives: using GP regression, weakly trained NNs, or Bayesian deep ensembles, where the latter two turn out to be equivalent to GP regression with particular covariance (kernel) functions.

\clearpage
\bibliography{references}

\begin{thebibliography}{14}
\providecommand{\natexlab}[1]{#1}
\providecommand{\url}[1]{\texttt{#1}}
\expandafter\ifx\csname urlstyle\endcsname\relax
  \providecommand{\doi}[1]{doi: #1}\else
  \providecommand{\doi}{doi: \begingroup \urlstyle{rm}\Url}\fi

\bibitem[Coker et~al.(2021)Coker, Pan, and Doshi-Velez]{coker2021wide}
Beau Coker, Weiwei Pan, and Finale Doshi-Velez.
\newblock Wide mean-field variational bayesian neural networks ignore the data.
\newblock \emph{arXiv preprint arXiv:2106.07052}, 2021.

\bibitem[He et~al.(2020)He, Lakshminarayanan, and Teh]{he2020bayesian}
Bobby He, Balaji Lakshminarayanan, and Yee~Whye Teh.
\newblock Bayesian deep ensembles via the neural tangent kernel.
\newblock \emph{arXiv preprint arXiv:2007.05864}, 2020.

\bibitem[Jacot et~al.(2018)Jacot, Gabriel, and Hongler]{jacot2018neural}
Arthur Jacot, Franck Gabriel, and Cl{\'e}ment Hongler.
\newblock Neural tangent kernel: Convergence and generalization in neural
  networks.
\newblock \emph{arXiv preprint arXiv:1806.07572}, 2018.

\bibitem[Kapoor et~al.(2010)Kapoor, Grauman, Urtasun, and
  Darrell]{kapoor2010gaussian}
Ashish Kapoor, Kristen Grauman, Raquel Urtasun, and Trevor Darrell.
\newblock Gaussian processes for object categorization.
\newblock \emph{International journal of computer vision}, 88\penalty0
  (2):\penalty0 169--188, 2010.

\bibitem[Kim and Lee(2007)]{kim2007clustering}
Hyun-Chul Kim and Jaewook Lee.
\newblock Clustering based on gaussian processes.
\newblock \emph{Neural computation}, 19\penalty0 (11):\penalty0 3088--3107,
  2007.

\bibitem[L{\'a}zaro-Gredilla and Figueiras-Vidal(2010)]{lazaro2010marginalized}
Miguel L{\'a}zaro-Gredilla and An{\'\i}bal~R Figueiras-Vidal.
\newblock Marginalized neural network mixtures for large-scale regression.
\newblock \emph{IEEE transactions on neural networks}, 21\penalty0
  (8):\penalty0 1345--1351, 2010.

\bibitem[LeCun et~al.(2015)LeCun, Bengio, and Hinton]{lecun2015deep}
Yann LeCun, Yoshua Bengio, and Geoffrey Hinton.
\newblock Deep learning.
\newblock \emph{nature}, 521\penalty0 (7553):\penalty0 436--444, 2015.

\bibitem[Lee et~al.(2019)Lee, Xiao, Schoenholz, Bahri, Novak, Sohl-Dickstein,
  and Pennington]{lee2019wide}
Jaehoon Lee, Lechao Xiao, Samuel Schoenholz, Yasaman Bahri, Roman Novak, Jascha
  Sohl-Dickstein, and Jeffrey Pennington.
\newblock Wide neural networks of any depth evolve as linear models under
  gradient descent.
\newblock \emph{Advances in neural information processing systems},
  32:\penalty0 8572--8583, 2019.

\bibitem[Neal(1996)]{neal1996priors}
Radford~M Neal.
\newblock Priors for infinite networks.
\newblock In \emph{Bayesian Learning for Neural Networks}, pages 29--53.
  Springer, 1996.

\bibitem[Snoek et~al.(2015)Snoek, Rippel, Swersky, Kiros, Satish, Sundaram,
  Patwary, Prabhat, and Adams]{snoek2015scalable}
Jasper Snoek, Oren Rippel, Kevin Swersky, Ryan Kiros, Nadathur Satish,
  Narayanan Sundaram, Mostofa Patwary, Mr~Prabhat, and Ryan Adams.
\newblock Scalable bayesian optimization using deep neural networks.
\newblock In \emph{International conference on machine learning}, pages
  2171--2180. PMLR, 2015.

\bibitem[Thakur et~al.(2021)Thakur, Lorsung, Yacoby, Doshi-Velez, and
  Pan]{thakur2021uncertaintyaware}
Sujay Thakur, Cooper Lorsung, Yaniv Yacoby, Finale Doshi-Velez, and Weiwei Pan.
\newblock Uncertainty-aware (una) bases for bayesian regression using
  multi-headed auxiliary networks, 2021.

\bibitem[Wang et~al.(2019)Wang, Pleiss, Gardner, Tyree, Weinberger, and
  Wilson]{wang2019exact}
Ke~Wang, Geoff Pleiss, Jacob Gardner, Stephen Tyree, Kilian~Q Weinberger, and
  Andrew~Gordon Wilson.
\newblock Exact gaussian processes on a million data points.
\newblock \emph{Advances in Neural Information Processing Systems},
  32:\penalty0 14648--14659, 2019.

\bibitem[Williams(1997)]{williams1997computing}
Christopher~KI Williams.
\newblock Computing with infinite networks.
\newblock \emph{Advances in neural information processing systems}, pages
  295--301, 1997.

\bibitem[Yang(2020)]{yang2020tensor}
Greg Yang.
\newblock Tensor programs ii: Neural tangent kernel for any architecture.
\newblock \emph{arXiv preprint arXiv:2006.14548}, 2020.

\end{thebibliography}
\clearpage

\appendix

\end{document}